\documentclass[conference]{IEEEtran}
\IEEEoverridecommandlockouts
\usepackage{cite}
\usepackage{amsmath,amssymb,amsfonts}
\usepackage{algorithmic}
\usepackage{graphicx}
\usepackage{textcomp}
\usepackage{xcolor}
\usepackage{booktabs}
\usepackage{multirow}
\usepackage{multicol}
\usepackage{stfloats}
\usepackage{hyperref}
\usepackage{float}

\def\BibTeX{{\rm B\kern-.05em{\sc i\kern-.025em b}\kern-.08em
    T\kern-.1667em\lower.7ex\hbox{E}\kern-.125emX}}
\begin{document}

\title{Scaling Laws of Graph Neural Networks for Atomistic Materials Modeling$^*$}

\author{\IEEEauthorblockN{Chaojian Li$^{4}$, Zhifan Ye$^{4}$, Massimiliano Lupo Pasini$^{1}$, Jong Youl Choi$^{2}$, Cheng Wan$^{4}$, \\ Yingyan (Celine) Lin$^{4}$, and Prasanna Balaprakash$^{3}$}
\IEEEauthorblockA{\textit{$^{1}$Computational Sciences and Engineering Division, Oak Ridge National Laboratory} \\
\textit{$^{2}$Computer Science and Mathematics Division, Oak Ridge National Laboratory} \\
\textit{$^{3}$Computing and Computational Sciences Directorate, Oak Ridge National Laboratory} \\
\textit{$^{4}$ Georgia Institute of Technology}\\
% City, Country \\
\{cli851, zye327, cwan39, celine.lin\}@gatech.edu, \{lupopasinim, choij, pbalapra\}@ornl.gov}
}

\maketitle
\begin{abstract}
Atomistic materials modeling is a critical task with wide-ranging applications, from drug discovery to materials science, where accurate predictions of the target material property can lead to significant advancements in scientific discovery. Graph Neural Networks (GNNs) represent the state-of-the-art approach for modeling atomistic material data thanks to their capacity to capture complex relational structures. While machine learning performance has historically improved with larger models and datasets, GNNs for atomistic materials modeling remain relatively small compared to large language models (LLMs), which leverage billions of parameters and terabyte-scale datasets to achieve remarkable performance in their respective domains.
To address this gap, we explore the scaling limits of GNNs for atomistic materials modeling by developing a foundational model with billions of parameters, trained on extensive datasets in terabyte-scale. Our approach incorporates techniques from LLM libraries to efficiently manage large-scale data and models, enabling both effective training and deployment of these large-scale GNN models. This work addresses three fundamental questions in scaling GNNs: the potential for scaling GNN model architectures, the effect of dataset size on model accuracy, and the applicability of LLM-inspired techniques to GNN architectures. Specifically, the outcomes of this study include (1) insights into the scaling laws for GNNs, highlighting the relationship between model size, dataset volume, and accuracy, (2) a foundational GNN model optimized for atomistic materials modeling, and (3) a GNN codebase enhanced with advanced LLM-based training techniques. Our findings lay the groundwork for large-scale GNNs with billions of parameters and terabyte-scale datasets, establishing a scalable pathway for future advancements in atomistic materials modeling.
\end{abstract}

\renewcommand\footnoterule{\noindent\rule{\linewidth}{1pt}\vspace{2pt}}
\renewcommand{\thefootnote}{$\ast$}
\footnotetext{\footnotesize \noindent This manuscript has been authored in part by UT-Battelle, LLC, under contract DE-AC05-00OR22725 with the US Department of Energy (DOE). The US government retains and the publisher, by accepting the article for publication, acknowledges that the US government retains a nonexclusive, paid-up, irrevocable, worldwide license to publish or reproduce the published form of this manuscript, or allow others to do so, for US government purposes. DOE will provide public access to these results of federally sponsored research in accordance with the DOE Public Access Plan (\url{http://energy.gov/downloads/doe-public-access-plan}).}

\section{Introduction}
\begin{figure}[t]
    \centering    \includegraphics[width=\linewidth]{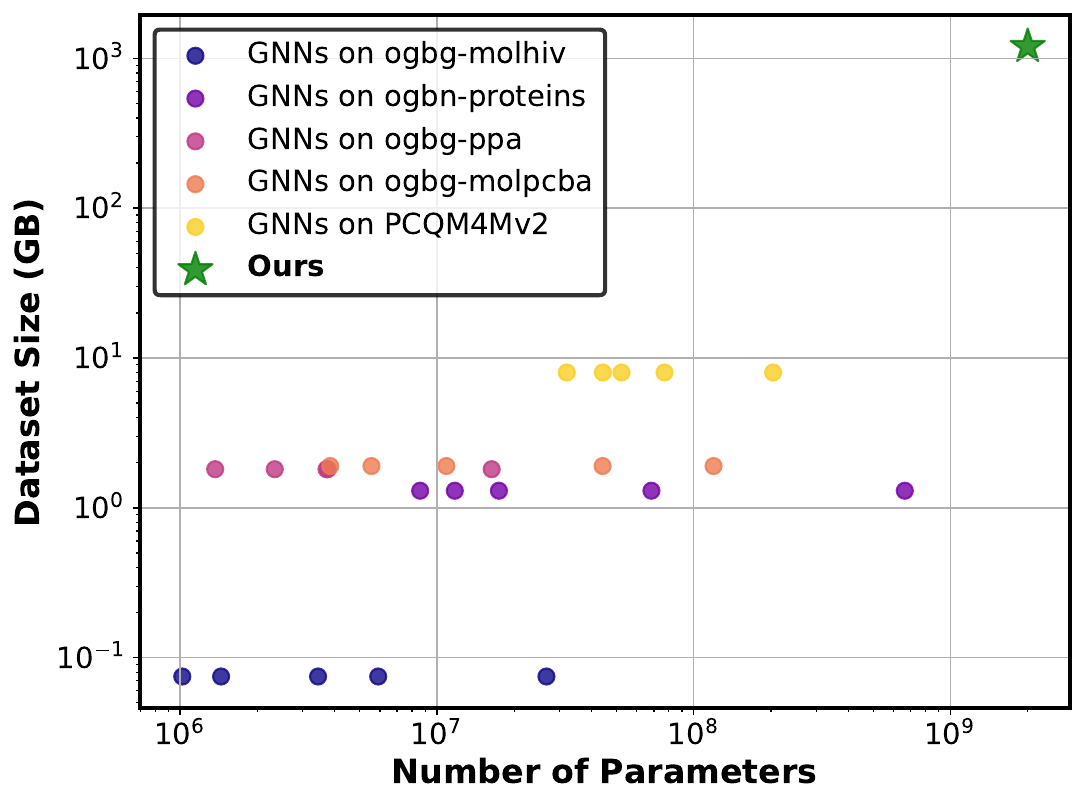}
    \vspace{-2em}
    \caption{Comparison of large-scale GNNs on multiple commonly-used biology/chemistry materials modeling datasets~\cite{hu2020open} with the foundational GNN developed in this work (indicated by a green star), after scaling both the model size and dataset size.}
    \vspace{-1em}
    \label{fig:teaser}
\end{figure}

Atomistic material modeling is essential for the discovery of new materials with target properties because it accelerates the process by predicting material properties with sufficiently accurate results using atomistic information, without requiring costly full first-principles calculations~\cite{pasini2024scalable,lupo2024first}. Neural networks play an important role in atomistic material modeling by training on data collected from chemistry experiments and first-principles calculations, formalizing material modeling as a neural network inference process~\cite{balabin2009neural}. Among these neural networks, Graph Neural Networks (GNNs)~\cite{kipf2016semi,velivckovic2017graph,satorras2021n} have emerged as the state-of-the-art (SOTA) approach because atomistic material structures can be naturally mapped onto a graph, where atoms and interatomic bonds are treated as the nodes and edges of a graph, respectively~\cite{sypetkowski2024scalability,chanussot2021open}.

However, the great promise of GNNs for atomistic material modeling has not yet been fully realized because the potential of larger GNN models and more extensive graph datasets has not been fully exploited~\cite{sypetkowski2024scalability,liu2024neural,liu2023towards}. Specifically, as highlighted in recent studies on language and image-based neural networks~\cite{kaplan2020scaling,zhai2022scaling}, larger neural networks trained on more comprehensive datasets tend to better capture the complex relationships in input data, leading to improved accuracy and reduced prediction errors. This scalability is often described by scaling laws, which indicate that increasing model size and dataset size generally improves prediction quality. In contrast, GNNs for atomistic material modeling currently lag behind in terms of the scale of models and datasets used compared to language and image-based neural networks. For example, as shown in Fig.~\ref{fig:teaser}, prior scaled-up GNNs are typically trained on datasets in the range of megabytes or gigabytes and have millions of parameters~\cite{pasini2024scalable,liu2024neural}, whereas commonly used language models today are trained on terabyte-scale datasets and contain billions of parameters~\cite{touvron2023llama}.

The aforementioned gap raises a series of fundamental questions about scaling up GNNs: (1) What is the potential of scaling GNN model architectures when the number of parameters reaches the billion level? (2) What is the effect of dataset size on achieved accuracy or prediction errors when scaled to the terabyte level? (3) Can prior techniques for scaling up distributed training of large-scale language and image-based neural networks also be applied to GNNs?

To address the questions raised above, we progressively scale up GNN model sizes from millions of parameters to billions, alongside a corresponding increase in the atomistic materials modeling dataset size to the terabyte level, as depicted in Fig.~\ref{fig:teaser}.
In summary, our contributions are threefold:

\begin{itemize}
    \item \textbf{Insight}: Extracted scaling laws for GNNs in atomistic materials modeling, offering guidance on the relationship between GNN model sizes, the amount of available training data, and the prediction error of material properties.
    \item \textbf{Model}: A foundational GNN model with billion-level parameters and terabyte-scale data, which, to the best of our knowledge, represents the largest GNN developed for atomistic materials modeling to date.
    \item \textbf{Infrastructure}: A scalable GNN training codebase integrated with SOTA distributed training techniques from language and image-based neural networks, enabling efficient training of billion-level GNNs with terabyte-scale atomistic materials modeling data.
\end{itemize}

\section{Related Works}

\subsection{GNN in Atomistic Materials Modeling}
GNNs have emerged as a powerful tool for atomistic materials modeling, offering state-of-the-art (SOTA) performance across various tasks in chemistry and materials science~\cite{rittig2023graph,buterez2024transfer}. These models leverage the inherent graph-like atomistic structures, where atoms are represented as nodes and chemical bonds as edges, allowing for a natural and intuitive representation of atomistic materials~\cite{wu2023chemistry}. GNNs excel in capturing complex relationships between atoms and their local chemical environments, enabling them to learn meaningful representations that can be used to predict a wide range of atomistic materials' properties~\cite{ma2022cross}. Recent advancements in GNN architectures have led to significant improvements in prediction accuracy, with models like EGNN~\cite{satorras2021n} achieving chemical accuracy on benchmark datasets such as QM9~\cite{reiser2022graph}. These models have demonstrated their ability to predict quantum mechanical properties, solubility, toxicity, and other crucial molecular characteristics with high precision. The success of GNNs in property prediction has opened up new possibilities for accelerating drug discovery, materials design, and other fields where understanding structure-property relationships is crucial. However, the challenge remains in scaling these models to handle even larger datasets and more complex atomistic structures, a gap this paper aims to address.

\subsection{GNN Foundation Model}
Graph Foundation Models (GFMs) represent a significant advancement in graph representation learning, designed to learn transferable representations that can generalize across diverse graph structures and tasks~\cite{liu2023towards}. These models aim to overcome the traditional limitations of task-specific GNNs by learning universal graph representations that can be applied to previously unseen graphs and tasks. Notable examples include GraphAny~\cite{zhao2024graphany}, which successfully demonstrates the ability to perform node classification across multiple graphs with varying feature dimensions and class numbers. In the realm of atomistic materials modeling, HydraGNN-GFM~\cite{pasini2024scalable} has emerged as a pioneering GFM architecture, specifically designed for large-scale scientific applications. HydraGNN-GFM's unique features include multi-task learning capabilities, flexible message passing neural network layers, and scalable distributed training support, allowing it to process hundreds of millions of graphs efficiently across thousands of GPUs. This scalability has been demonstrated through near-linear strong scaling performance on major supercomputing systems, processing over 154 million atomistic structures while maintaining high prediction accuracy. Motivated by the scalability demonstrated by HydraGNN-GFM, we adopt the same GNN model architecture for our scaling experiments in this work.

\subsection{Scalable Training Techniques}
Scalable training techniques are crucial for enabling GNNs to handle large-scale graphs that exceed the memory capacity of a single machine. Traditional distributed training approaches, widely adopted in deep learning~\cite{verbraeken2020survey}, primarily rely on data parallelism, which distributes data batches across multiple devices while maintaining full model replicas on each device, synchronizing gradients during updates. Optimizer-level parallelism methods such as ZeRO~\cite{rajbhandari2020zero}, further reduce memory requirements by distributing optimizer states across devices. Conversely, model parallelism partitions the model architecture itself across devices, necessitating careful management of intermediate activations. Another memory-saving strategy, activation checkpointing, lowers memory usage by rematerializing intermediate activations during backward propagation~\cite{chen2016training}. For GNNs, distributed training introduces unique challenges due to the interconnected nature of graph data. Partition-based methods, which divide the graph into subgraphs distributed across devices, mitigate these challenges by distributing computational workloads but often incur significant communication overhead from exchanging neighbor features~\cite{zhang2023survey}.

\section{Infrastructure Setup}

\begin{figure}[t]
    \centering    \includegraphics[width=\linewidth]{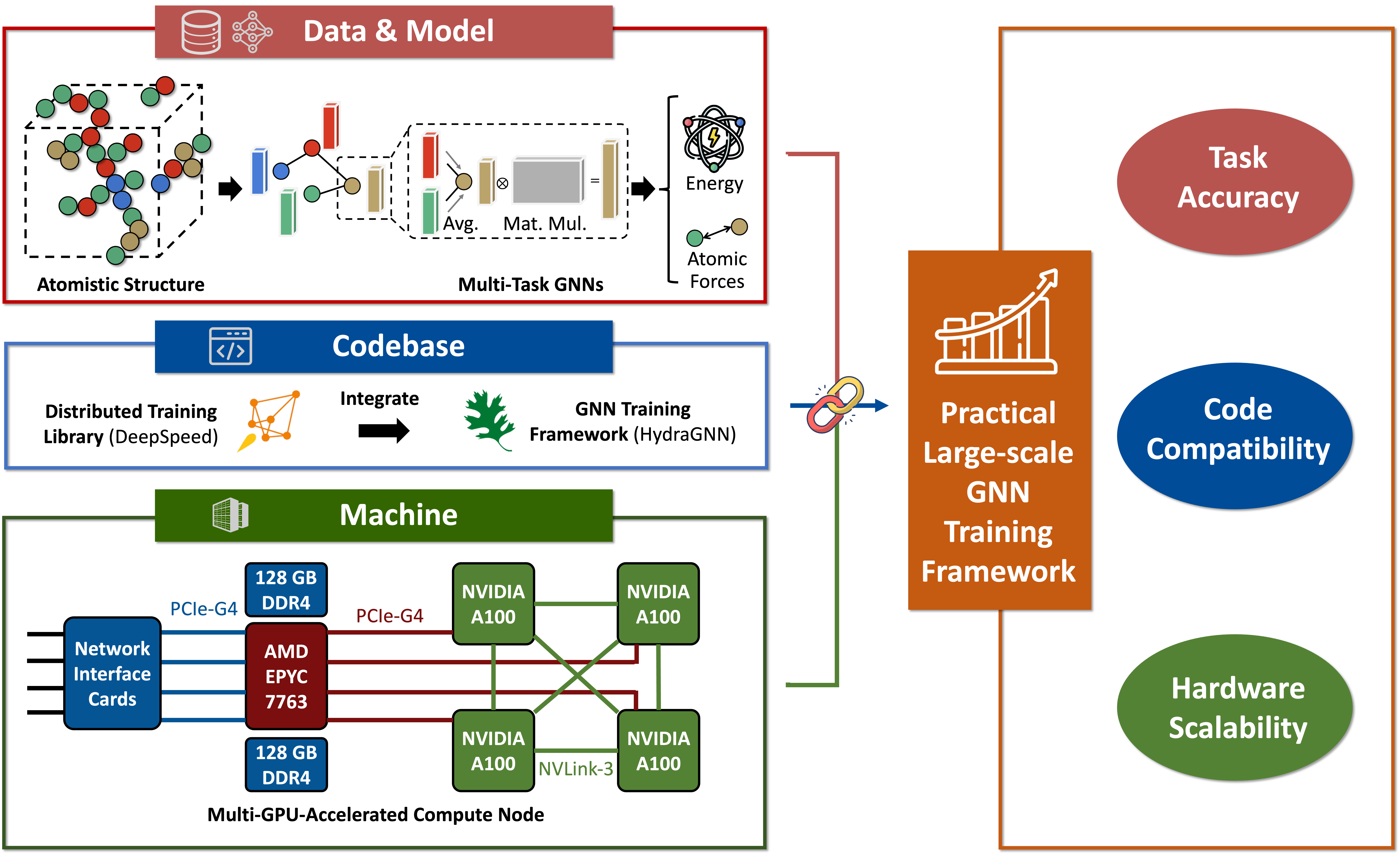}
    \vspace{-0.5em}
    \caption{An overview of the developed multi-stack infrastructure for scalable GNN training, integrating data and model configuration, refactored codebase, and multi-GPU machine setup. This unified framework aims to simultaneously improve GNN task accuracy through data-driven model design, ensure code modularity and reuse via codebase restructuring, and optimize training efficiency and scalability with a multi-GPU hardware architecture.} 
    \label{fig:setup}
\end{figure}

As summarized in Fig.~\ref{fig:setup}, the entire infrastructure used for the scaling experiments includes not only the data and model but also the codebase supporting scalable GNN training and the high-performance computing (HPC) cluster.

\subsection{Data}
The dataset used in our scaling experiments is aggregated from multiple publicly available atomistic materials modeling datasets. As summarized in Tab.~\ref{tab:data}, each dataset contains a different number of samples(graphs). Specifically, ANI1x~\cite{smith2020ani} consists of up to 57 thousand distinct molecular configurations featuring the chemical elements C, H, N, and O. QM7-X~\cite{hoja2021qm7} includes 42 physicochemical properties for approximately 4.2 million equilibrium and non-equilibrium structures of small organic molecules. OC2020-20M~\cite{chanussot2021open} is a subsampled set from the original OC2020~\cite{chanussot2021open} dataset, spanning a range of oxide materials, coverages, and adsorbates. OC2022~\cite{tran2023open} is an updated version of OC2020~\cite{chanussot2021open}, containing additional samples. MPTrj~\cite{jain2013commentary} focuses on atomistic structures of inorganic materials. Combining the five aforementioned data sources resulted in an aggregated dataset of 1.2 TB. We further sampled multiple datasets with sizes ranging from 0.1 TB to 1.2 TB from the original 1.2 TB dataset and used them in the scaling experiments.
Following~\cite{pasini2024scalable,liu2024neural}, the corresponding tasks in the aggregated dataset are defined as predicting energy and atomic forces from atomistic structures represented in graph format. Specifically, the energy to be predicted is a property of the entire atomistic structure (graph), whereas the atomic forces are properties of individual atoms.

\begin{table}[!b]\centering
\vspace{-1em}
\caption{Summary of the data sources of the aggregated dataset used in the scaling experiments.}
\scriptsize
\setlength{\tabcolsep}{2pt}
\resizebox{\linewidth}{!}{\begin{tabular}{c||c|c|c|c}\toprule
Data Source  & \# of Nodes & \# of Edges & \# of Graphs & Size \\
\midrule
ANI1x~\cite{smith2020ani} & 75,700,481 & 1,050,357,960 & 4,956,005 & 25 GB \\
QM7-X~\cite{hoja2021qm7} & 70,675,659 & 1,020,408,506 & 4,195,237 & 25 GB \\
OC2020-20M~\cite{chanussot2021open} & 1,538,055,547 & 33,734,466,610 & 20,994,999 & 726 GB \\
OC2022~\cite{tran2023open} & 705,379,388 & 18,937,505,384 & 8,834,760 & 395 GB \\
MPTrj~\cite{jain2013commentary} & 49,286,440 & 729,940,098 & 1,580,227 & 17 GB \\
\bottomrule
\end{tabular}
}
\label{tab:data}
\end{table}

\subsection{Model}
To ensure equivalence with rotations, translations, reflections, and permutations in atomistic materials modeling, we select EGNN~\cite{satorras2021n} as the GNN model type, which is specifically designed for predicting molecular properties. To align with the task definitions in the aggregated dataset used for our scaling experiments, we add two types of output heads on top of the EGNN models: one for graph-level property prediction and the other for node-level property prediction. To study the effectiveness of the backbone model independently of task types when scaling data and model sizes, we vary only the depth and width of the EGNN backbone during the scaling experiments. The hyperparameter settings used for training follow those in~\cite{pasini2024scalable}, e.g., all models are trained for 10 epochs.

\subsection{Codebase}
We conduct the scaling experiments using HydraGNN~\cite{HydraGNN}, a commonly used GNN training framework for scientific discovery. However, as discussed in Sec.~\ref{sec:training_techniques}, while HydraGNN supports data parallelism, directly scaling the model and data to sizes beyond those previously encountered in the codebase can exceed memory limits, even on high-performance HPC machines equipped with A100 GPUs. To address this limitation, we integrate DeepSpeed, a well-known distributed learning library, into HydraGNN to alleviate memory constraints, enabling us to scale our models to billions of parameters and handle terabyte-level datasets. The corresponding modifications have been committed to the official HydraGNN repository at: \url{https://github.com/ORNL/HydraGNN}.

\subsection{Machine}
We utilize Perlmutter~\cite{nersc_perlmutter}, an HPC cluster consisting of A100-accelerated compute nodes. Each node is equipped with an AMD EPYC 7763 CPU, 256 GB of DDR4 memory, and four NVIDIA A100 GPUs interconnected via NVLink-3. Additionally, to accelerate the data loading process on the HPC machines, we employ the ADIOS~\cite{godoy2020adios} scientific data management library and DDStore~\cite{choi2023ddstore}, a distributed data store that facilitates in-memory data transfer between processes. In particular, we use 32 compute nodes for training each model.

\section{Scaling Laws in GNN}
To examine the scaling laws for GNNs in atomistic material modeling, we conducted experiments on scaling up GNN model sizes (Sec.~\ref{sec:model_scaling}), increasing dataset size (Sec.~\ref{sec:data_scaling}), and analyzing how model depth and width affect the final test loss (Sec.~\ref{sec:depth-vs-width}). Specifically, model scaling is achieved by increasing the number of neurons in each layer, while dataset scaling is accomplished by adding samples to the training set from the aggregated 1.2 TB dataset. Additionally, the test loss for different model and dataset combinations is evaluated using the same held-out test set from the 1.2 TB dataset.

\subsection{Model Scaling}
\label{sec:model_scaling}
\begin{figure*}[t]
    \centering
    \vspace{-1em}
    \includegraphics[width=\linewidth]{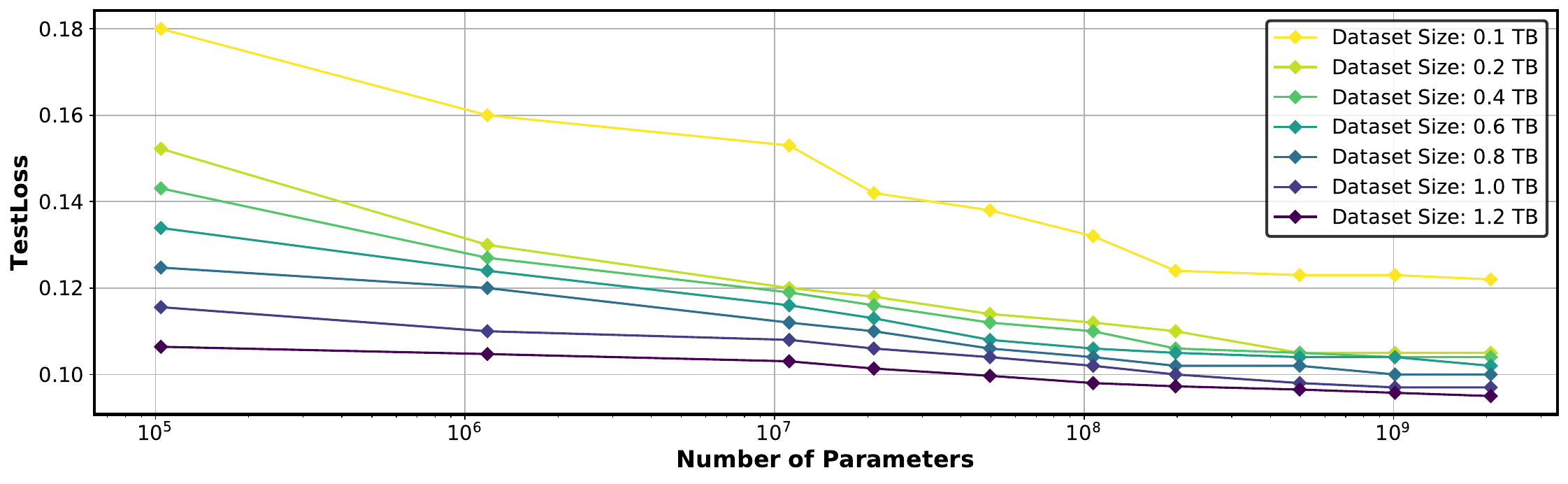}
    \vspace{-2em}
    \caption{The effect of scaling GNN \textbf{model} sizes across various dataset sizes on the final test loss.}
    \label{fig:model-scaling}
\end{figure*}

\begin{figure*}[b]
    \centering  
    \vspace{-1em}
    \includegraphics[width=\linewidth]{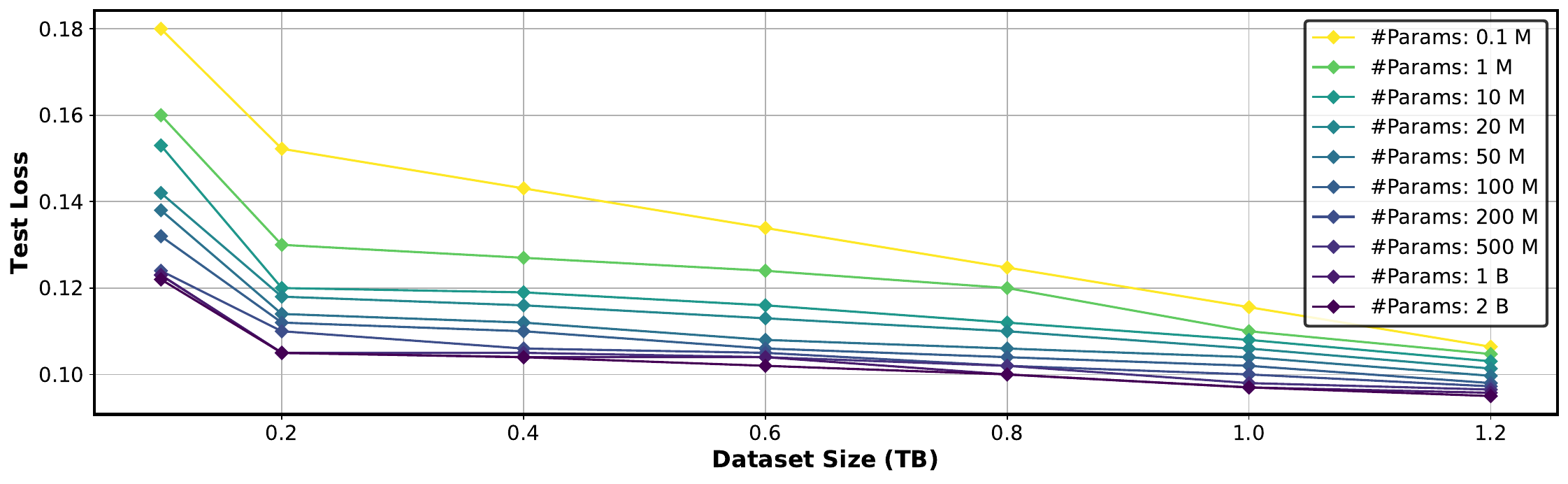}
    \vspace{-2em}
    \caption{The effect of scaling atomistic materials modeling \textbf{dataset} sizes across various GNN model sizes on the final test loss.}
    \label{fig:data-scaling}
\end{figure*}

As shown in Fig.~\ref{fig:model-scaling}, when the model size is scaled up by increasing the number of neurons in each layer, the test loss consistently decreases. This trend is observed across datasets of different sizes, ranging from 0.1 TB to 1.2 TB. However, unlike the observations in large-scale language and image-based models~\cite{zhai2022scaling,kaplan2020scaling}, where the loss is nearly linear with the log-scale of the number of parameters, the decrease in test loss for GNNs in atomistic materials modeling shows diminishing returns as model size increases. We conjecture that this is due to architectural differences between Transformer models~\cite{vaswani2017attention}, commonly used in language and image-based tasks, and GNNs used in atomistic materials modeling. Specifically, Transformer models~\cite{vaswani2017attention} rely on attention mechanisms, which can adaptively learn connections between different input samples (tokens). In contrast, GNN architectures, even advanced ones like EGNN~\cite{satorras2021n} that account for rotational, translational, reflective, and permutational equivalences, are inherently limited by their locality constraints. These constraints restrict the ability to freely learn connections between any pair of nodes, reducing their overall learning capacity compared to Transformers.

Although there has been some exploration of applying Transformers to graphs or encoding graphs directly into Transformer tokens~\cite{vaswani2017attention,yun2019graph}, their effectiveness in atomistic materials modeling remains underexplored. Furthermore, many of these approaches are limited to solving simple graph statistics-related problems~\cite{wei2023unleashing,guo2023gpt4graph}.

In conclusion, \textbf{further scaling of GNN model sizes is a promising direction for improving the quality of atomistic materials modeling}. However, when scaling beyond 2 billion parameters, the limitations of current GNN architectures may become a bottleneck.

\subsection{Data Scaling}
\label{sec:data_scaling}
As summarized in Fig.~\ref{fig:data-scaling}, the test loss of GNNs with varying parameter counts, ranging from 0.1 million to 2 billion, consistently decreases as more data becomes available for training. In particular, when the dataset size increases from 0.1 TB to 0.2 TB, there is a noticeable drop in test loss compared to the more gradual decreasing trend observed beyond 0.2 TB. We conjecture that this is due to the significant differences between the 0.1 TB subset and the full 1.2 TB dataset. The test set is sampled from the complete 1.2 TB dataset and remains fixed across all experiments. As a result, when only 0.1 TB is sampled for training, there is likely a mismatch between the distribution of the training dataset and the test set, leading to a relatively higher test loss for the 0.1 TB setting.

In contrast, as the dataset size gradually increases from 0.2 TB to 1.2 TB, the test loss decreases steadily and predictably, consistent with observations in existing large-scale language and image-based models~\cite{zhai2022scaling,kaplan2020scaling}. The more pronounced loss reduction, even at the 1.2 TB scale, suggests that \textbf{scaling data is more effective than scaling model size when both reach relatively large scales}, i.e., millions of parameters and terabytes of data.

However, it is important to note that scaling data is more challenging than scaling models. Simply increasing model depth and width produces a larger model, whereas scaling data involves tedious processes such as data collection, cleaning, and verification, as demonstrated in~\cite{pasini2024scalable}. Therefore, the bottleneck in exploring the limits of scaling GNN models for atomistic materials modeling is similar to the bottleneck faced in scaling language and image-based models: the need for large-scale, high-quality data.

\subsection{Model Depth vs. Width}
\label{sec:depth-vs-width}

\begin{figure}[t]
    \centering    \includegraphics[width=\linewidth]{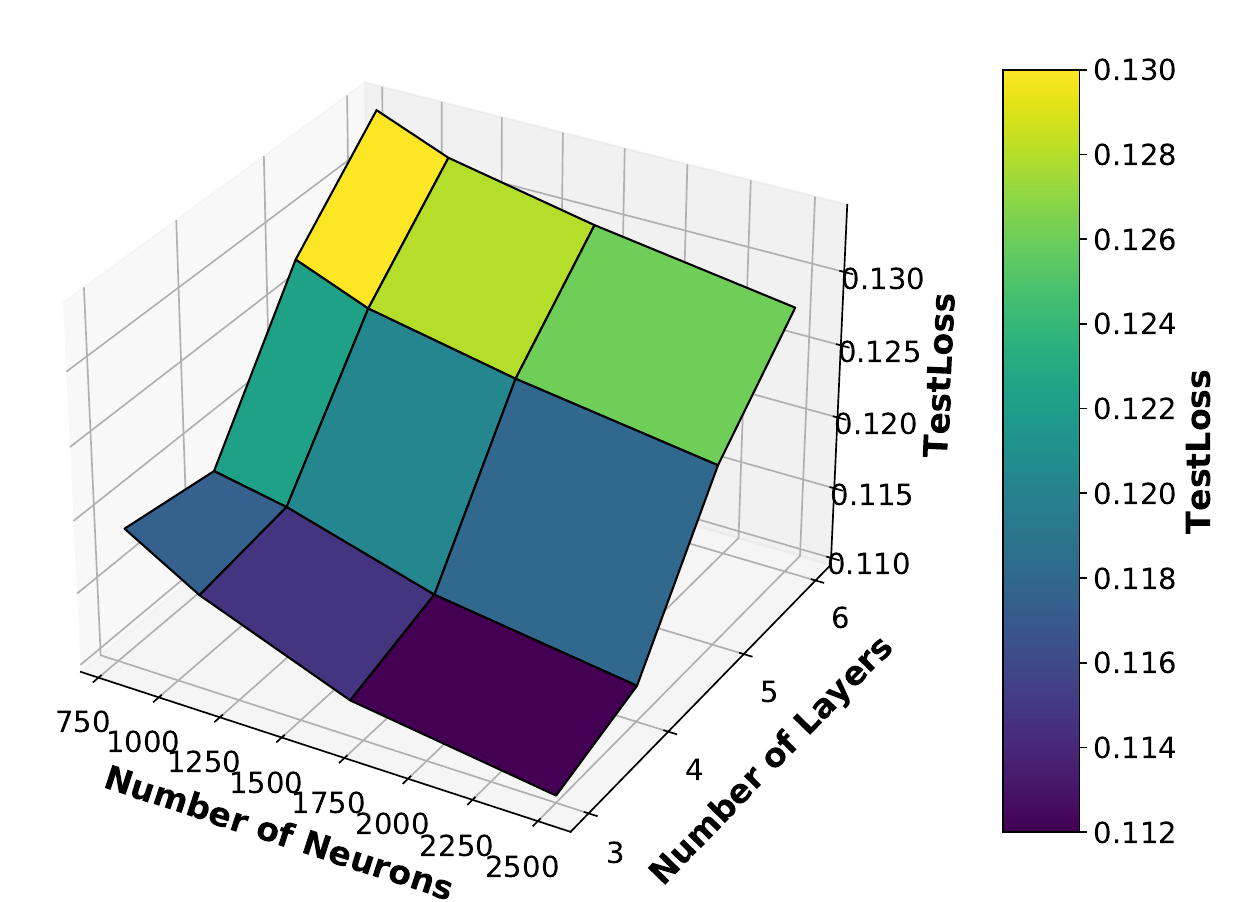}
    \vspace{-1em}
    \caption{Comparison of how scaling GNN model \textbf{depth} (i.e., number of layers) and \textbf{width} (i.e., number of neurons in each layer) affects the test loss when training on a dataset of 0.4 TB in size.}
    \label{fig:depth-vs-width}
\end{figure}

As noted in previous works on scaling neural networks, a model's depth and width are two critical factors that influence the achieved test loss~\cite{zhai2022scaling,kaplan2020scaling}. Examining the effect of depth and width in GNN models is particularly important, as it is widely recognized that building deep GNNs is more challenging compared to convolutional neural networks or Transformer models~\cite{li2019deepgcns,li2020deepergcn}. Motivated by prior observations on scaling depth and width, we conducted experiments to explore how these two factors affect relatively large-scale GNN models with 10 to 100 million parameters on a substantial 0.4 TB dataset.

As summarized in Fig.~\ref{fig:depth-vs-width}, our findings indicate that increasing model width—i.e., the number of neurons in each layer—consistently results in lower test loss. In contrast, increasing the number of layers beyond three leads to higher test loss, even when the total model size increases. We hypothesize that the over-smoothing~\cite{chen2020measuring} issue inherent to GNN architectures persists even at such large-scale dataset sizes and model capacities. This suggests that scaling GNN models by \textbf{increasing the number of neurons per layer is a more effective strategy than increasing the number of layers}.

\section{Effectiveness of Training Techniques}
To support training models with millions of parameters and datasets at terabyte scales, using the pure PyTorch-based data-distributed training in HydraGNN leads to out-of-memory issues, even on mainstream A100 GPUs. To address this, we investigated (1) the peak memory bottlenecks in large-scale GNN training (Sec.~\ref{sec:bottleneck}) and (2) whether commonly used methods for scalability in large language models can be effectively applied to GNNs in atomistic material modeling (Sec.~\ref{sec:ckpt} and Sec.~\ref{sec:zero}).

\label{sec:training_techniques}
\subsection{Bottleneck Profiling}
\label{sec:bottleneck}

\begin{figure}[t]
    \centering  
    % \vspace{-1em}
    \includegraphics[width=\linewidth]{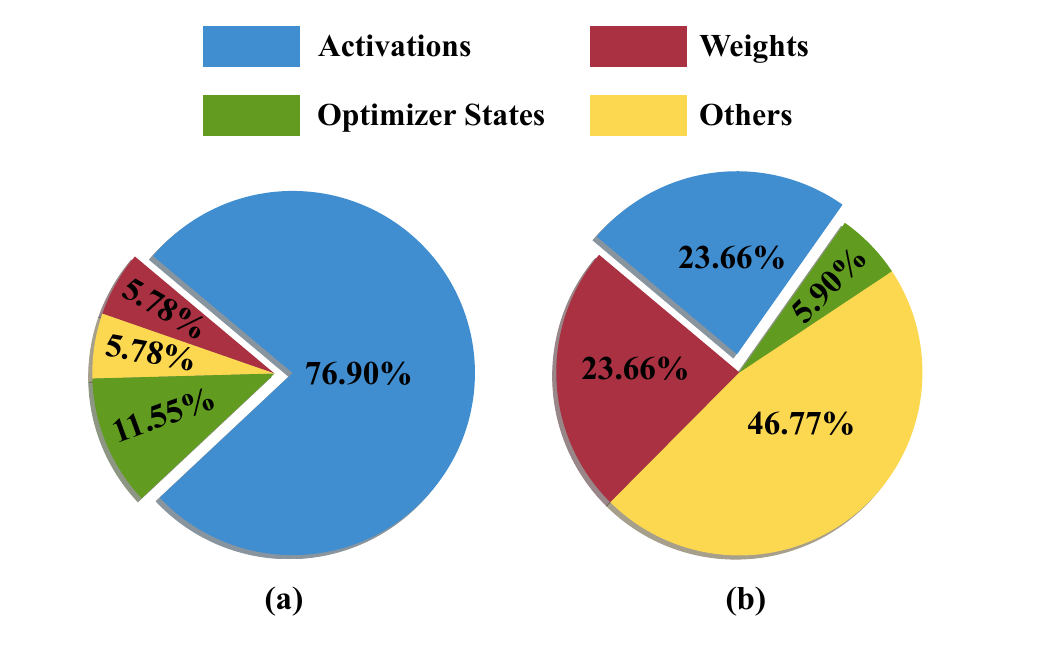}
    \vspace{-2em}
    \caption{Summary of the peak memory usage breakdown when training GNNs using (a) vanilla PyTorch-based HydraGNN~\cite{HydraGNN} and (b) HydraGNN integrated with activation checkpointing and the ZeRO optimizer~\cite{rajbhandari2020zero}.}
    % \vspace{-0.5em}
    \label{fig:profile}
\end{figure}

Limited memory capacity has long been a critical bottleneck in training large deep learning models~\cite{rajbhandari2021zero}. To overcome this limitation, numerous memory-efficient training techniques have been proposed~\cite{chen2016training, duan2024efficient}, primarily targeting applications in image and text domains. However, the memory constraints in training large GNNs pose similar challenges, limiting their scalability~\cite{wan2022bns}. To better understand and address these challenges, we conducted a memory bottleneck profiling analysis to evaluate the memory consumption in training foundational GNNs and to identify effective techniques for scalable GNN training. Since memory usage varies throughout the training process, our analysis focuses on profiling \textit{peak memory usage}, defined as the highest memory consumption observed during training. This peak typically occurs across three stages: (1) the forward pass, (2) the backward pass, and (3) the weight updates performed by the optimizer.

Fig.~\ref{fig:profile} (a) illustrates the peak memory usage breakdown during GNN training without any memory-efficient optimizations. The results indicate that the peak memory usage arises at the start of the backward pass. The breakdown highlights two key insights: \underline{(1)} \textbf{Activations} dominate peak memory usage, accounting for 76.90\% of the total. These tensors store intermediate values computed during the forward pass and are essential for gradient computation during the backward pass; \underline{(2)} \textbf{Optimizer states} represent the second largest contributor to peak memory usage. These states, maintained by the Adam optimizer~\cite{kingma2014adam}, include momentum vectors, which require storage equivalent to twice the size of the model weights.

\subsection{Activation Checkpointing}
\label{sec:ckpt}

To mitigate the substantial memory overhead caused by activation tensors, we implemented the activation checkpointing technique~\cite{chen2016training} in HydraGNN framework~\cite{HydraGNN}. This method reduces memory consumption by selectively recomputing partial activations during the backward pass, rather than storing all intermediate activations generated during the forward pass. By applying this technique, activation tensors are no longer the dominant contributor to memory usage, resulting in a significant 58$\%$ reduction in peak memory usage. Following this optimization, the new peak memory usage shifts to the weight update phase. However, as shown in Tab.~\ref{tab:training_techniques}, activation checkpointing comes at the cost of a 10$\%$ increase in training latency due to the overhead introduced by recomputation.

\subsection{ZeRO Optimizer}
\label{sec:zero}

After adopting activation checkpointing, optimizer states became the largest contributor to peak memory usage. To address this, we incorporated the ZeRO optimizer~\cite{rajbhandari2020zero} by adding DeepSpeed library~\cite{rasley2020deepspeed} in HydraGNN framework~\cite{HydraGNN}, which reduces memory demands by partitioning optimizer states across multiple GPUs instead of duplicating them on each device. With four GPUs within one compute node, this approach achieved a 36$\%$ reduction in peak memory usage compared to using activation checkpointing alone, as shown in Fig.~\ref{fig:profile}. However, as elaborated in Tab.~\ref{tab:training_techniques}, this optimization introduces additional cross-GPU communication overhead, leading to a 133\% increased training runtime.

\begin{table}[t]\centering
\caption{Reduction in peak memory usage and the overhead of training time after adopting activation checkpointing and ZeRO Optimizer.}
\scriptsize
\setlength{\tabcolsep}{2pt}
\resizebox{\linewidth}{!}{\begin{tabular}{l||c|c}\toprule
\textbf{Setting} & \textbf{Relative Peak Memory} & \textbf{Relative Training Time}  \\
\midrule
Vanilla Pytorch & 100$\%$ & 100$\%$ \\
+ Activation Checkpointing & 42$\%$ & 110$\%$ \\
+ ZeRO Optimizer & 27$\%$ & 133$\%$ \\
\bottomrule
\end{tabular}
\vspace{-1em}
}
\label{tab:training_techniques}
\end{table}

\section{Impact of the Scaled Models and the Infrastructure}

As demonstrated in Sec.~\ref{sec:model_scaling}, the scaled-up GNN models demonstrate improved accuracy over smaller versions. This improvement highlights the potential of scaling in addressing the complexities of molecular property prediction tasks. The enhanced performance of these models could advance a range of scientific applications. For example, in material discovery, these models can enable rapid exploration of vast compositional and structural spaces, predicting key properties such as mechanical strength, thermal conductivity, and electronic behavior. By predicting critical properties, this capability accelerates the identification of novel materials for energy storage, catalysis, and sustainable manufacturing. Similarly, in drug design, the enhanced predictive power of these models can facilitate the identification of drug candidates with desired pharmacological properties, streamlining the traditionally time-consuming and resource-intensive process of lead optimization. Beyond these applications, the techniques and practices of the delivered infrastructure lay a foundation for scaling GNN models further and adapting them to broader applications in molecular science and other domains including biology, sociology, and more.

\section{Conclusion}
This work bridges the gap between current GNNs for atomistic materials modeling and advances in scalable training techniques and scaling laws observed in large language and image-based models. By scaling GNNs to billions of parameters and terabyte-level datasets, we achieve significant improvements in predicting material properties. Our findings uncover scaling laws for GNNs, emphasizing the relationship between model size, dataset volume, and prediction accuracy, and establish a new benchmark in this field. The extracted insights, large-scale foundation models, and infrastructure developed in this work enable efficient handling of large-scale models and data, advancing atomistic materials modeling and expanding scientific applications.

\section*{Acknowledgements}
Massimiliano Lupo Pasini would like to thank Dr. Vladimir Protopopescu for his valuable feedback in the preparation of the manuscript.
This research is sponsored by the Artificial Intelligence Initiative as part of the Laboratory Directed Research and Development (LDRD) Program of Oak Ridge National Laboratory, managed by UT-Battelle, LLC, for the US Department of Energy under contract DE-AC05-00OR22725.
This work used resources of the Oak Ridge Leadership Computing Facility, which is supported by the Office of Science of the U.S. Department of Energy, under INCITE award CPH161. This work also used resources of the National Energy Research
Scientific Computing Center, which is supported by the Office of Science of the U.S. Department of Energy under Contract No. DE-AC02-05CH11231, under awards ERCAP0027259 and ERCAP0030519.
This work was supported in part by the National Science Foundation (NSF) (Award ID: 2400511 and 2016727), the Department of Health and Human Services Advanced Research Projects Agency for Health (ARPA-H) under Award Number AY1AX000003, and CoCoSys, one of the seven centers in JUMP 2.0, a Semiconductor Research Corporation (SRC) program sponsored by DARPA.

\bibliographystyle{IEEEtranS}
\bibliography{refs}

\end{document}